\documentclass{article} 
\usepackage{iclr2027_conference,times}

\usepackage{amsmath,amsfonts,bm}

\def\eqref#1{equation~\ref{#1}}

\def\1{\bm{1}}

\DeclareMathAlphabet{\mathsfit}{\encodingdefault}{\sfdefault}{m}{sl}
\SetMathAlphabet{\mathsfit}{bold}{\encodingdefault}{\sfdefault}{bx}{n}

\usepackage{hyperref}
\usepackage{graphicx}
\usepackage{url}
\usepackage{booktabs}
\usepackage{xcolor}
\usepackage{booktabs, multirow}
\usepackage{colortbl}
\usepackage{float}
\usepackage{listings}
\usepackage{booktabs}

\lstdefinestyle{prompt}{
  basicstyle=\scriptsize\ttfamily,
  breaklines=true,
  breakatwhitespace=true,   
  columns=fullflexible,     
  keepspaces=true,
  breakindent=0pt,          
  postbreak={},             
}

\title{Scaling 3D Visual Grounding in Abdominal CT}

\author{%
  Sam Church$^{1,}$\thanks{\begin{tabular}[t]{@{}l@{}}
    Corresponding author: \href{mailto:sdchurch@wisc.edu}{\texttt{sdchurch@wisc.edu}} \\
    Code: \href{https://github.com/samdchurch/LocusCT}{\texttt{github.com/samdchurch/LocusCT}}
  \end{tabular}},
  Danyal Maqbool$^{1}$,
  Joshua D. Warner$^{2}$,
  Andrew Voter$^{2}$, \\
  \textbf{Junjie Hu}$^{1}$,
  \textbf{Meghan G. Lubner}$^{2}$,
  \textbf{Tyler J. Bradshaw}$^{2}$ \\
  \\
$^{1}$\,University of Wisconsin--Madison, Department of Computer Sciences, Madison, WI USA \\
$^{2}$\,University of Wisconsin--Madison, Department of Radiology, Madison, WI USA}
\iclrfinalcopy 
\begin{document}

\maketitle

\fancyhead{}                          
\renewcommand{\headrulewidth}{0pt}    

\begin{abstract}

Visual grounding models can enhance radiology workflows by linking report findings to image regions. This is particularly valuable for 3D CT, where findings often occupy a tiny fraction of the volume. Training 3D grounding models requires large sets of paired phrases and regions, and building such datasets is expensive, requiring radiologists to annotate images by hand. We posit that this supervision is already created implicitly during routine reporting, as radiologists frequently place 2D annotations (e.g., distance measurement, arrows) on key images to make measurements and to support report interpretation. We introduce an automated pipeline that converts these routine clinical annotations into large-scale phrase–region supervision for 3D visual grounding. The pipeline links each annotation to the corresponding finding in the report through metadata matching, then uses a promptable 3D segmentation model to convert the 2D annotation into a volumetric mask. This produces phrase–mask–volume datasets without requiring additional radiologist annotation. Applied to a single institution’s clinical picture archiving and communication system (PACS), our approach generated 105K phrase–mask–volume triplets from 59K abdominal CT exams. We also introduce two abdominal CT grounding benchmarks, LocusBench-Onc and LocusBench-ED, which comprise 240 oncology and 260 emergency-department radiologist-reviewed phrase–mask–volume triplets, respectively, with the latter spanning 13 distinct categories such as appendicitis, hematoma, and hernia. We further introduce LocusCT, a 3D visual grounding model trained on this dataset, which achieves hit rates of 0.725 on LocusBench-Onc and 0.773 on LocusBench-ED, substantially outperforming comparator models. These results show that routine PACS annotations are a scalable, previously unused source of supervision for 3D visual grounding.

\end{abstract}


\section{Introduction}
\label{sec:intro}

\begin{figure}
    \centering
    \includegraphics[width=\linewidth]{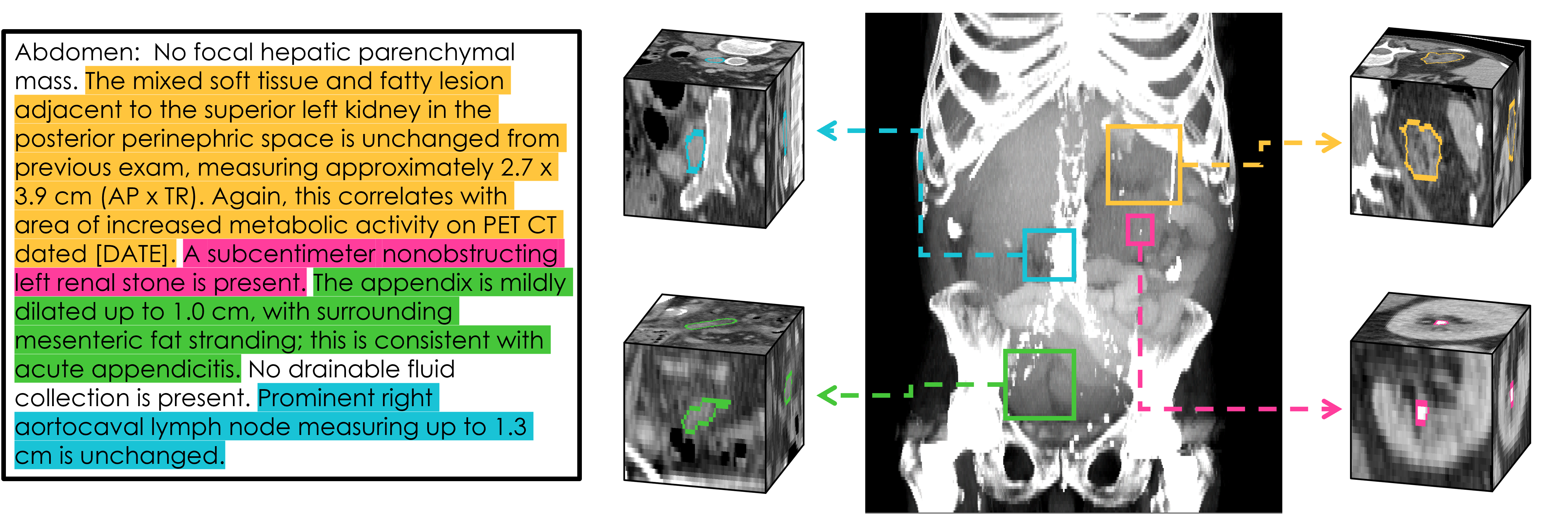}
    \caption{\textbf{Visual grounding of clinical findings in 3D abdominal CT.} A radiology report contains multiple findings referring to distinct anatomical regions within the CT volume. Colored phrases in the report correspond to their respective localized regions in the 3D abdominal CT, illustrated by the matching colored boxes and volume crops.}
    \label{fig:task_figure}
\end{figure}

Visual grounding is the task of localizing the image region referred to by a free-text phrase. When applied to radiology, it could lead to more efficient workflows and enhanced reporting capabilities, particularly for computed tomography (CT). During routine clinical care, radiologists interpret CT volumes and dictate findings into free-text reports. The findings correspond to features within the images (e.g., “3.2 cm hypoattenuating lesion in hepatic segment VII”), yet in current workflows, the text and corresponding image locations are not linked. Automatically linking these (Fig.~\ref{fig:task_figure}) could enable rapid visual verification of reported findings, facilitate comparison across follow-up exams, and connect radiology reports to image regions for downstream vision-language systems.

While grounding has advanced rapidly in natural images \citep{refcoco, mdetr, grounding_dino} and, more recently, in 2D medical imaging such as chest radiography \citep{MS_CXR, MIMO, medmo}, progress in CT has been considerably slower. This is for two main reasons. First, multimodal labeled data is scarce: voxel-level annotation of findings in CT volumes is time-consuming and requires expert radiologists, so existing 3D datasets either cover anatomical structures rather than pathology \citep{total_segmentator} or contain a limited number of finding-level annotations, largely restricted to the chest \citep{rexgroundingct}. Second, the task itself is difficult: findings are often very small relative to the scan volume, vary widely in appearance, and are described in reports in varying levels of detail using heterogeneous language.

To address the data scarcity problem, we observe that a vast and largely unused source of localization supervision already exists in clinical archives. During routine interpretation, radiologists place measurement lines, arrows, and region markers on the images. These markers are stored in the picture archiving and communication system (PACS) using a standardized file format: DICOM grayscale softcopy presentation state (GSPS) objects. Moreover, because radiologists choose what to mark, the annotations tend to target the findings they consider most clinically important.

\begin{figure}
    \centering
    \includegraphics[width=\linewidth]{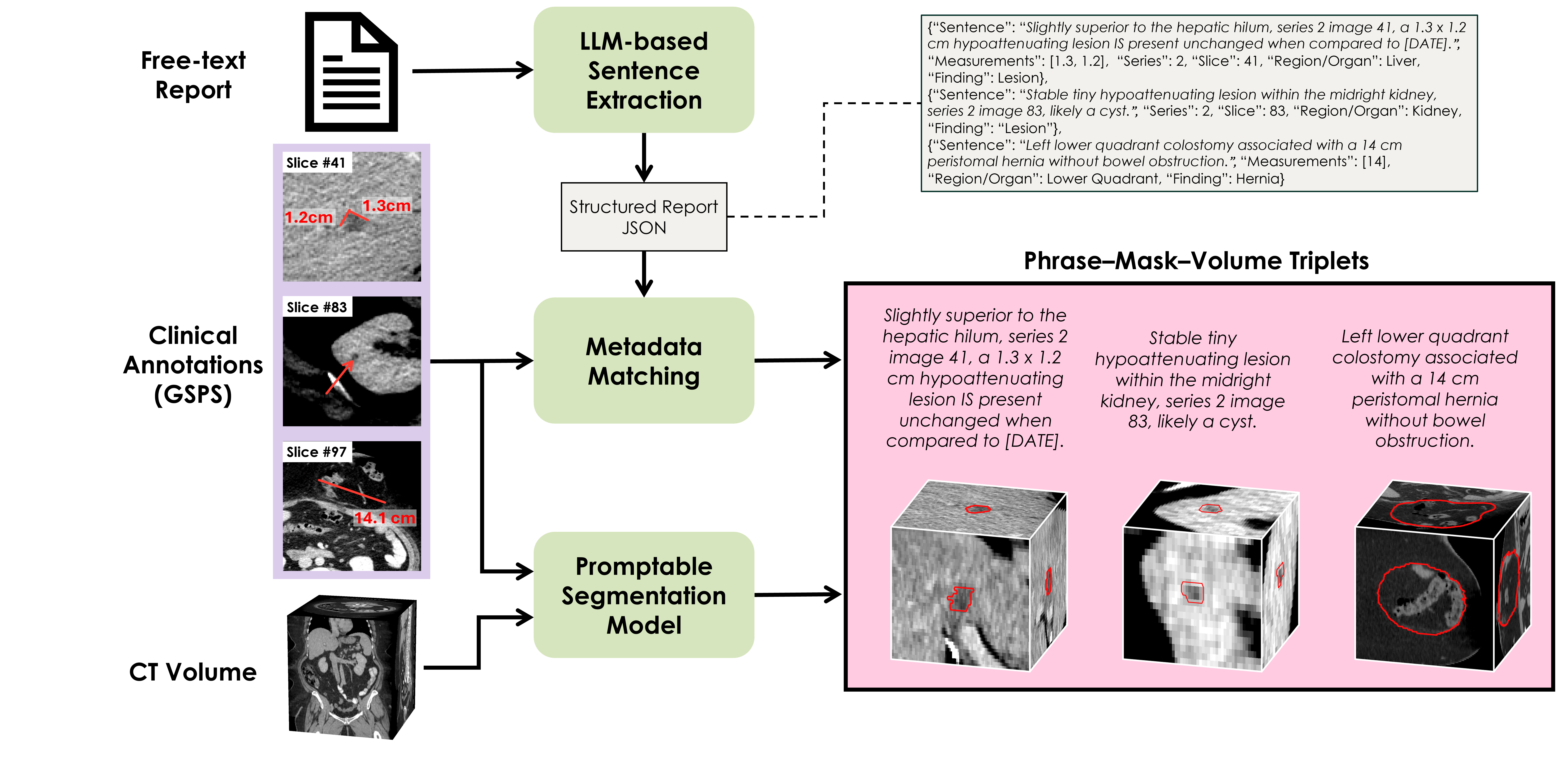}
    \caption{\textbf{Automated generation of phrase–mask–volume triplets from routine clinical data.} An LLM extracts finding information from free-text reports, which are matched to 2D GSPS annotations using exam metadata. A promptable 3D segmentation model converts the matched 2D annotation into a volumetric mask, yielding grounding supervision without additional radiologist annotation.}
    \label{fig:pipeline_figure}
\end{figure}

In this work, we introduce a new pipeline for generating visual grounding training data from these routine annotations (Fig.~\ref{fig:pipeline_figure}). We extract GSPS objects from abdominal CT examinations at our institution, convert each 2D mark into a 3D segmentation using SAM2CT \citep{SAM2CT}, and pair each resulting region with its corresponding report sentence using GSPS metadata. SAM2CT is a promptable segmentation model built on SAM2 \citep{SAM2} and designed specifically to convert GSPS annotations into 3D CT segmentations. This process yields 105K finding--region pairs from 59K studies without any dedicated annotation effort.

To enable visual grounding evaluation, we introduce LocusBench, a public benchmark for abdominal CT grounding. Two board-certified radiologists reviewed and corrected 500 examples, which we release as two expert-verified test sets: LocusBench-Onc (N=240), consisting of findings suggestive of malignancy sampled without predefined categories, and LocusBench-ED (N=260), sampled evenly across 13 finding categories common in emergency imaging, such as appendicitis, diverticulitis, abscess, and aneurysm. We evaluate current 3D grounding and text-promptable segmentation models on LocusBench and find that none perform well on abdominal findings. This suggests that capabilities learned from existing grounding datasets do not transfer to this setting.

Finally, we use our weakly labeled dataset to train visual grounding models. We compare two architectures, ConTEXTual Net 3D \citep{contextualNet3D} and VoxTell \citep{voxtell}, and build LocusCT, a visual grounding model for abdominal CT findings based on VoxTell. LocusCT achieves strong performance on both oncology and emergency-department abdominal CT findings, demonstrating that weak supervision derived from routine clinical annotations can support effective 3D visual grounding.

Our key contributions are:
\begin{itemize}
\item \textbf{A scalable weak-labeling pipeline.} We convert routine PACS annotations into 105K phrase--mask--volume triplets for abdominal CT without any radiologist annotation.

\item \textbf{LocusBench.} We introduce an expert-verified benchmark for abdominal CT grounding with 500 cases spanning oncology findings and 13 emergency-department finding categories. Existing grounding models achieve hit rates below 0.25 on these benchmarks.

\item \textbf{LocusCT.} We train a 3D grounding model on our weakly labeled dataset, achieving hit rates of 0.725 and 0.773 on LocusBench-Onc and LocusBench-ED, respectively, more than tripling the strongest comparator. Evaluation on the external Merlin dataset provides initial evidence of transfer beyond our institution.
\end{itemize}

\section{Related Work}
\label{sec:related}
 
\paragraph{Visual grounding.}
Visual grounding (also known as referring expression comprehension) localizes the image region described by a natural-language phrase. Early approaches formulated referring expression comprehension as ranking candidate region proposals according to their correspondence with a referring expression \citep{mattnet}, with benchmarks such as RefCOCO providing standardized evaluation \citep{refcoco}. Subsequent transformer-based methods, including MDETR \citep{mdetr} and Grounding DINO \citep{grounding_dino}, unified language understanding with end-to-end localization, while recent vision-language models extend grounding to conversational and open-ended prompts~\citep{kosmos-2, shikra, qwen3-vl}. Despite this progress, most visual grounding datasets rely on manually annotated phrase–region pairs, which are particularly scarce for 3D medical imaging.

\paragraph{Visual grounding in medical imaging.} Medical visual grounding has been studied primarily in 2D chest radiography, where datasets such as MS-CXR~\citep{MS_CXR}, ChEX~\citep{ChEX}, and uMedGround~\citep{uMedGround} link report language to localized findings. More recently, text-promptable 3D medical segmentation models, including SegVol~\citep{segvol}, SAT~\citep{SAT}, ConTEXTual Net 3D~\citep{contextualNet3D}, BiomedParse~\citep{biomedparse}, MedVL-SAM2~\citep{medvlsam2}, and VividMed~\citep{vivid_med}, have enabled language-guided localization and segmentation across volumetric data. However, these models are largely trained on public segmentation datasets where text prompts correspond to anatomical structures or predefined disease categories, rather than fine-grained clinical findings. Two recent works are most closely related to ours. ReXGroundingCT~\citep{rexgroundingct} provides findings-level grounding data from 3,142 non-contrast chest CT scans, with grounding annotations focused on lung and pleural abnormalities. Its reports originate from Turkish clinical reports that were machine-translated and subsequently rewritten with GPT-4 for standardization. VoxTell~\citep{voxtell} incorporates ReXGroundingCT alongside other segmentation datasets, accepts free-text prompts, and is trained on over 62K volumes spanning more than 1K anatomical and pathological classes. In contrast, we train and evaluate on a broad range of abnormal findings in both contrast-enhanced and non-contrast abdominal CT, deriving supervision directly from radiologists' routine clinical annotations while retaining their original report text. This yields heterogeneous phrase--mask--volume pairs that reflect clinical reporting language and scale to 105K annotated findings without additional manual labeling.

\paragraph{Mining clinical annotations from PACS.}
Routine annotations created during clinical image interpretation provide an alternative source of supervision. DeepLesion~\citep{deeplesion} mined RECIST measurements from PACS to construct a large-scale CT lesion dataset, while subsequent work has mined physician annotations for MRI tumor detection~\citep{mri_tumor_detection_mined_annotations}. SAM2CT~\citep{SAM2CT} further showed that line and arrow annotations stored as DICOM GSPS objects can be converted into 3D segmentation masks using promptable segmentation. We extend this direction by linking GSPS annotations to the corresponding report findings and pairing them with volumetric masks, yielding phrase–mask–volume supervision for 3D visual grounding at scale.

\begin{figure}
    \centering
    \includegraphics[width=\linewidth]{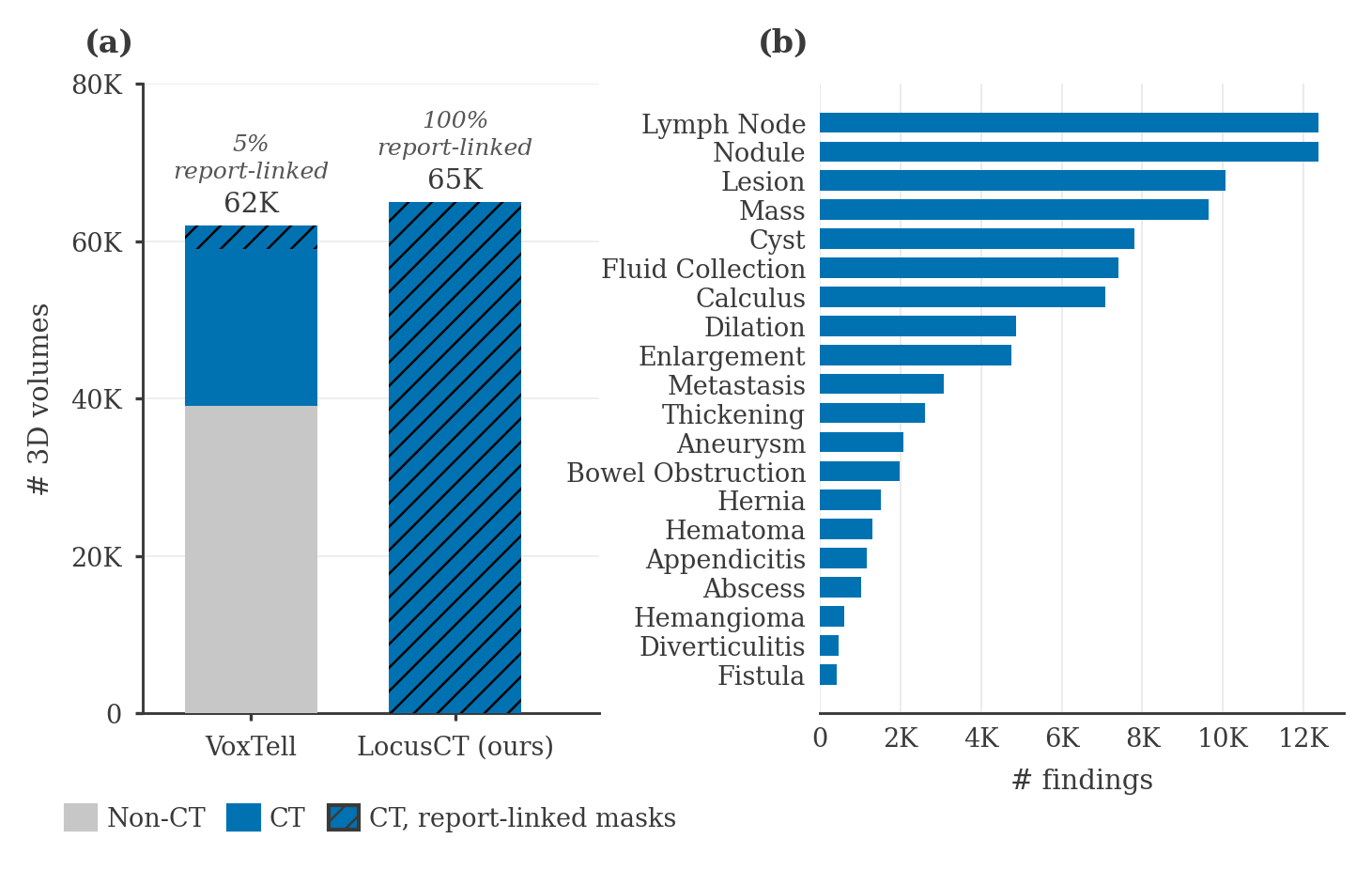}
    \caption{\textbf{Our method provides large-scale, report-linked supervision for 3D CT grounding.} \textbf{(a)} Composition of the training data for VoxTell and LocusCT. We call a volume report-linked if its segmentation masks are paired with finding phrases from the exam's own clinical report. Only 5\% of VoxTell's volumes (13\% of its CT volumes) are report-linked, whereas all 65K volumes (from 59K exams) in LocusCT's training set are. \textbf{(b)} Distribution of the 105K findings across the 20 most frequent categories, which together cover 88.1\% of all findings.}
    \label{fig:data_stats_fig}
\end{figure}

\section{Weak-Labeling Pipeline}
 \label{sec:weak_labeling}
\subsection{Background: GSPS Annotations}
 
\paragraph{Annotations in routine practice.}
During clinical interpretation, radiologists often annotate images directly within the PACS viewer, though annotation practices are not standardized and vary across institutions and individual radiologists. When retained, annotations are stored as DICOM GSPS objects, which encode the graphical marks and their spatial locations in the source image's coordinate system. Common annotation types include arrows for localizing findings, lines for measuring and localizing findings, ellipses that can summarize image statistics such as mean, minimum, and maximum Hounsfield units within a region, and free-text annotations. GSPS objects also record the corresponding series, image slice, and slice position, allowing each annotation to be mapped back to its location in the original CT examination.

Because these annotations are generated as part of routine clinical care rather than an independent labeling process, they can provide a large-scale source of clinically grounded supervision at institutions where annotation is common practice. However, they are not directly suitable for visual grounding: annotations are two-dimensional, do not explicitly identify the finding being marked, vary in whether and how findings are annotated, and are not linked to the report text describing the corresponding finding. Our pipeline addresses these limitations to convert routine clinical annotations into phrase--mask--volume triplets for training 3D visual grounding models.

\subsection{Data Collection}
We pulled every abdominal CT taken from our institution from a 10-year period. This resulted in a total of 91K exams from 3 main cohorts: oncology exams ordered by our institution's cancer center (26K exams), emergency department exams (31K exams), and all other available abdominal CTs from other parts of the hospital and outpatient clinics (34K exams). Using the pipeline detailed below and shown in Fig.~\ref{fig:pipeline_figure}, we converted these exams into a total of 105K phrase--mask--volume triplets, coming from a total of 65K unique volumes, 59K exams, and 37K unique patients. Fig.~\ref{fig:data_stats_fig}a compares the size and composition of our dataset with VoxTell's training data~\citep{voxtell}, and Fig.~\ref{fig:data_stats_fig}b shows the distribution of findings produced by our pipeline.

\subsection{Matching GSPS annotations to report phrases}
\label{sec:matching}

\paragraph{Sentence extraction.} We use Qwen3.8-27B \citep{qwen38} to extract report sentences that contain exactly one non-negative finding. For each extracted sentence, the model also returns the information needed for matching: series and slice number, measurement values, and slice position (SP). Of the 91K reports, 14K did not have matchable information. Of the remaining 77K reports, we extracted 200K sentences containing one or more of our matching criteria, with 61K containing series/slice references, 172K containing measurement values, and 15K containing an SP value. 

\paragraph{Line measurement matching.} We match each line measurement to a sentence reporting the same measurement. We allow for rounding or truncation when the match is unambiguous. For example, a 1.1~cm line measurement matches the sentence "There is a 1 cm cyst in the left kidney'' if no other line measurement in the exam also rounds to 1~cm. Two lines that intersect at near 90-degree angles are treated as a single bidimensional annotation and matched to sentences containing bidimensional or tridimensional measurements (e.g., "1.2 $\times$ 3.4 cm''). Among the extracted GSPS objects, we identified 117K unidimensional and 34K bidimensional line measurements. Of the 172K sentences containing measurement information, our pipeline matched 72.9K to unidimensional annotations and 24.7K to bidimensional annotations.

\paragraph{Arrow annotation matching.} We match arrows using either series/slice number pairs or slice position (SP). If a sentence references a specific slice and exactly one arrow annotation lies on that slice, we consider the two a match. From our 84K exams with GSPS annotations, we had 112K arrow annotations. From the 66K sentences containing series/slice references or SP values, we were able to match 30.0K arrow annotations.

\paragraph{Other GSPS annotations.} Our PACS supports additional GSPS annotation types, such as ellipses, which we describe in more detail in the Appendix. We exclude these annotations from the current pipeline and leave their integration for future work.

\subsection{Converting GSPS Annotations to 3D Segmentations}
\label{sec:segmentation}
GSPS annotations can serve to localize findings, but they do not capture the finding's boundaries. To close this gap, we use SAM2CT, a promptable segmentation model designed to convert 2D annotations into 3D segmentations. SAM2CT accepts the standard SAM2 \citep{SAM2} prompts (points, bounding boxes, and masks) as well as GSPS-style annotations (arrows, lines, and bidimensional lines). Following the SAM2 paradigm, it segments the annotated slice first and then propagates the segmentation slice by slice in both directions. Section~\ref{sec:locusbench} contains results on the accuracy of this 3D segmentation method within our pipeline.

\section{LocusBench}
\label{sec:locusbench}
Because no public benchmark exists for grounding abdominal CT findings, we developed LocusBench, a radiologist-verified benchmark with two test sets. LocusBench-Onc contains 240 cases randomly sampled from the oncology cohort, restricted to findings suggestive of cancer (e.g., tumor, mass, or lesion). LocusBench-ED contains 260 cases, 20 randomly sampled from each of 13 ED finding categories.

All LocusBench cases were first segmented using our weak-labeling pipeline and then reviewed by radiologists, who evaluated both of its components: sentence-to-GSPS matching and 3D segmentation quality. For matching, radiologists verified that each sentence corresponded to the GSPS object the algorithm had assigned to it, and that the referring phrase contained enough anatomical information to identify the finding. For segmentation, radiologists could reject a 3D mask for undersegmentation (defined as capturing less than approximately 75\% of the described finding), or for oversegmentation (defined as the finding occupying less than 50\% of the segmented volume). Radiologists also rated each mask on a five-point scale (1: complete failure; 2: major edits required; 3: moderate edits required; 4: minor edits required; 5: radiologist quality). Cases with a failed match, undersegmentation, or oversegmentation were revised by an experienced CT annotator and returned to the radiologists for further review. This was repeated until every case was approved.

Because the first-pass review was conducted on unedited pipeline outputs, it also serves as an evaluation of our weak-labeling pipeline. On this initial pass, the weakly labeled segmentations received mean scores of 4.27 on LocusBench-Onc and 4.32 on LocusBench-ED (out of 5). For LocusBench-Onc, radiologists flagged 7.5\% of cases as needing text edits and 11.7\% as needing segmentation corrections; for LocusBench-ED, the corresponding rates were 2.3\% and 16.9\%.

\begin{table}[t]
\caption{Comparison of 3D visual grounding models under zero-shot settings.}
\label{tab:model-comparisons}
\centering
\small
\setlength{\tabcolsep}{6pt}
\renewcommand{\arraystretch}{1.15}
\begin{tabular}{lcccc}
\toprule
& \multicolumn{2}{c}{\textbf{LocusBench-Onc}} 
& \multicolumn{2}{c}{\textbf{LocusBench-ED}} \\
\cmidrule(lr){2-3} \cmidrule(lr){4-5}
\textbf{Model} 
& \textbf{Dice} & \textbf{Hit Rate}
& \textbf{Dice} & \textbf{Hit Rate} \\

\midrule
\quad VoxTell        & 0.117 & 0.188 & 0.118 & 0.209 \\
\quad SegVol         &  0.015  &   0.044  &  0.016   &   0.037  \\
\quad BiomedParse-v2 & 0.065 & 0.158 & 0.067 & 0.182 \\
\bottomrule
\end{tabular}
\end{table}

\section{Experiments}

\subsection{Zero-shot evaluations}
SegVol \citep{segvol} and BiomedParse-v2 \citep{biomedparse} segment from text prompts, but are trained on anatomical category names rather than radiology report language. VoxTell \citep{voxtell} was trained with some clinical report text from chest CTs \citep{rexgroundingct}, but not abdominal CTs. Following prior work on report-grounded CT segmentation \citep{rexgroundingct}, we report 3D Dice and Hit Rate (the fraction of cases with $\text{Dice} \ge 0.1$). As shown in Table~\ref{tab:model-comparisons}, VoxTell performs the best of the comparator models with hit rates of 0.188 and 0.209 on the oncology and emergency department evaluations, respectively.

\begin{table}[t]

\centering
\caption{\textbf{Dice and hit rate on LocusBench.} Per-category results on LocusBench-ED are sorted by LocusCT$_{\text{VoxTell}}$ Dice. Diverticulitis is the most challenging category for both models, likely because its margins blend into adjacent inflammation. Aneurysm, cyst, and mass achieve the highest Dice.}
\label{tab:results}
\small
\setlength{\tabcolsep}{7pt}
\begin{tabular}{l cc cc}
\toprule
 & \multicolumn{2}{c}{LocusCT$_{\text{ConTEXTualNet}}$} & \multicolumn{2}{c}{LocusCT$_{\text{VoxTell}}$} \\
\cmidrule(lr){2-3} \cmidrule(lr){4-5}
 & Dice & Hit Rate & Dice & Hit Rate \\
\midrule
\rowcolor{gray!12} LocusBench-Onc & 0.462 & 0.596 & 0.495 & 0.725 \\
\rowcolor{gray!12} LocusBench-ED  & 0.468 & 0.731 & 0.490 & 0.773 \\
\midrule
\multicolumn{5}{l}{\textit{LocusBench-ED by category}} \\
\addlinespace[2pt]
\quad Cyst           & 0.735 & 0.950 & 0.685 & 0.950 \\
\quad Aneurysm       & 0.608 & 0.900 & 0.654 & 0.900 \\
\quad Mass           & 0.589 & 0.750 & 0.632 & 0.850 \\
\quad Lymph Node     & 0.403 & 0.800 & 0.548 & 0.800 \\
\quad Kidney Stone   & 0.617 & 0.900 & 0.537 & 0.850 \\
\quad Hematoma       & 0.583 & 0.700 & 0.511 & 0.850 \\
\quad Lesion         & 0.431 & 0.750 & 0.475 & 0.850 \\
\quad Hernia         & 0.344 & 0.550 & 0.459 & 0.650 \\
\quad Appendicitis   & 0.302 & 0.700 & 0.445 & 0.850 \\
\quad Nodule         & 0.557 & 0.850 & 0.431 & 0.700 \\
\quad Gallstone      & 0.368 & 0.700 & 0.419 & 0.750 \\
\quad Abscess        & 0.354 & 0.600 & 0.377 & 0.700 \\
\quad Diverticulitis & 0.193 & 0.350 & 0.202 & 0.350 \\
\bottomrule
\end{tabular}
\end{table}

\subsection{Model Training}
\label{sec:training}
To evaluate the efficacy of our weak-labeling process, we trained two models on architectures designed for grounding free-form clinical text: ConTEXTual Net 3D and VoxTell. Both are built on a 3D nnU-Net backbone and fuse text features progressively in the decoder. ConTEXTual Net 3D fuses them with cross-attention, whereas VoxTell uses a simpler scheme: it multiplies the text and image features and concatenates the result with the image features.

For both models, CT volumes were resampled to $1.5 \times 1.5 \times 3.0$~mm and cropped to $352 \times 352 \times 192$ voxels. ConTEXTual Net 3D processes each cropped volume in a single pass, whereas VoxTell operates on $192 \times 192 \times 192$ patches and uses sliding-window tiling at inference. Following \cite{pillar-0}, each volume is given to the model as a 3-channel input, with the CT windowed to soft-tissue, lung, and bone settings. Both models were initialized from scratch and trained on 4 NVIDIA H200 GPUs until convergence. The VoxTell-based model performed best, and we release it as LocusCT. 

\begin{figure}
    \centering
    \includegraphics[width=\linewidth]{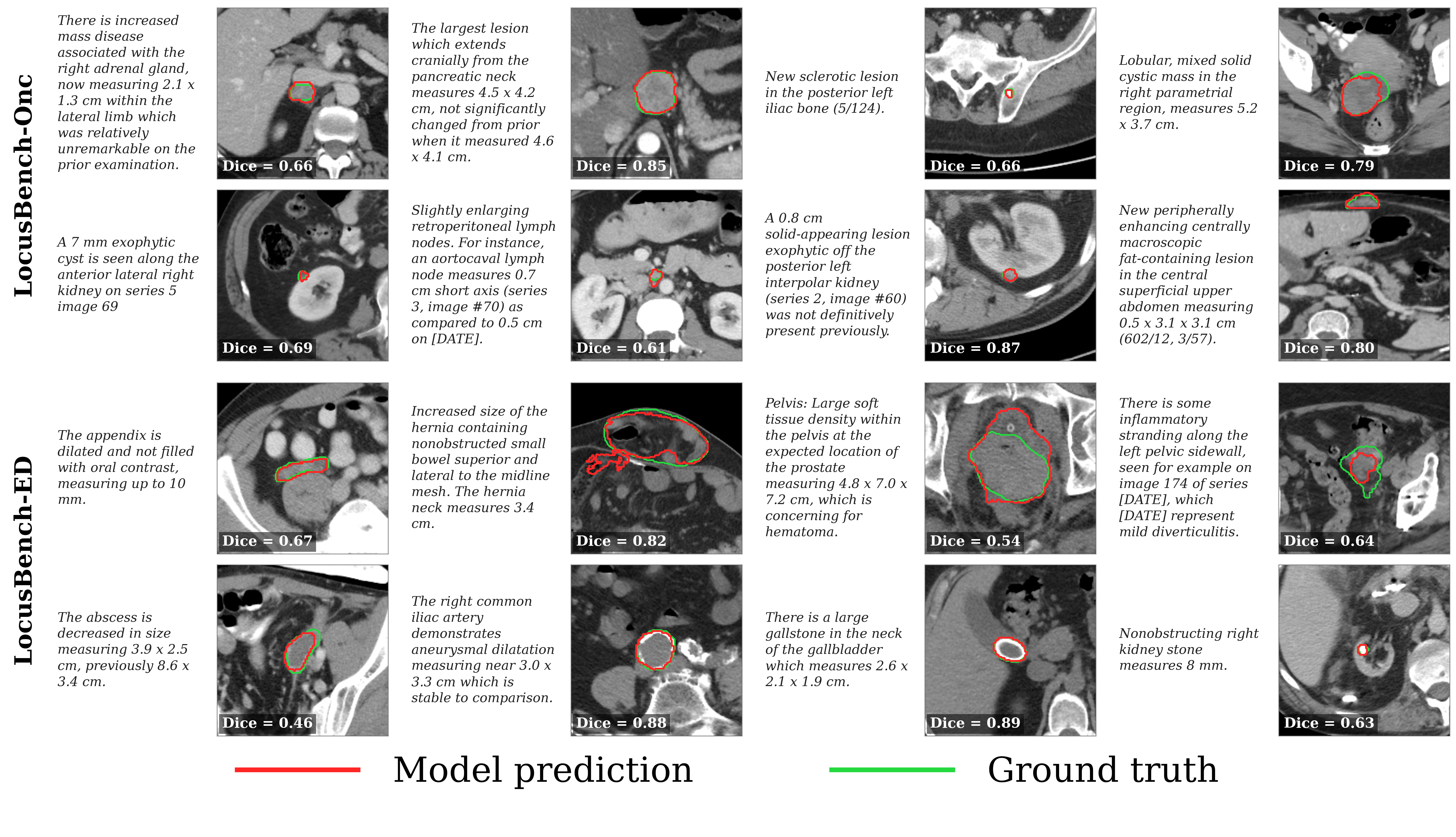}
    \caption{\textbf{LocusCT predictions on LocusBench-Onc and LocusBench-ED.} Each example shows a clinical finding description alongside a representative CT slice with the model prediction (red), radiologist-annotated ground truth (green), and Dice score.}
    \label{fig:locusbench_examples}
\end{figure}

LocusCT achieved an average hit rate of 0.725 and 0.773 on LocusBench-Onc and LocusBench-ED, respectively. Detailed results for both models are shown in Table~\ref{tab:results}. Visualization of LocusCT's performance on LocusBench-Onc and LocusBench-ED can be seen in Fig.~\ref{fig:locusbench_examples}. 

\subsection{Ablations and Analysis}

We perform several ablations to assess how training data scale, input representation, and supervision curation affect LocusCT performance. First, to evaluate the benefit of our large-scale training set, we train models using 10\% and 30\% of the 105K phrase--mask--volume triplets and compare them with the full-data model. This measures whether performance continues to benefit from scaling.

Second, we investigate the effect of the input representation by training a model using a single input channel rather than the three-channel representation used by LocusCT. 

Finally, we examine whether the automatically generated supervision should be restricted to findings resembling those used to train SAM2CT. We construct a \emph{curated} training set by retaining only samples whose findings fall within the distribution represented in SAM2CT's training data, and compare this model with one trained on all available matched GSPS annotations. This experiment tests whether the model benefits from a larger but less reliable training dataset, versus a smaller but more precise training dataset. Results for all ablations are reported in Table~\ref{tab:ablations}.

\subsection{Merlin: External Evaluation}
\label{sec:merlin}
\begin{figure}
    \centering
    \includegraphics[width=\linewidth]{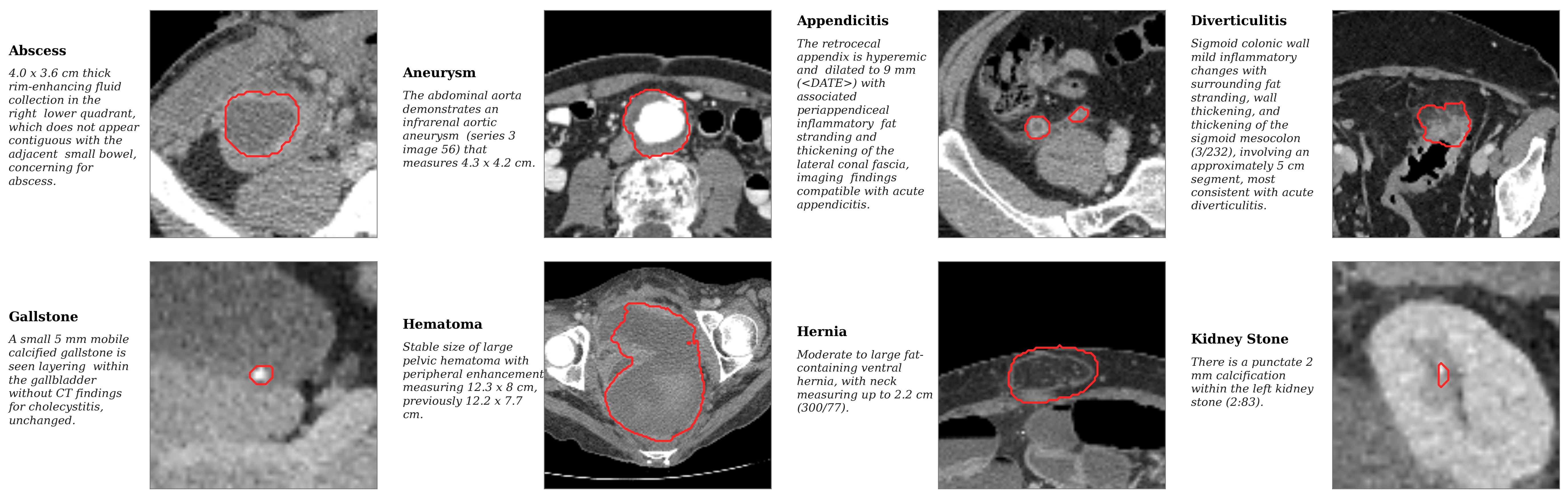}
    \caption{\textbf{LocusCT predictions on the external Merlin dataset.} Report phrases (left) and predicted masks in red (right), across eight ED categories. Merlin has no mask annotations, so these examples are qualitative only.}
    \label{fig:merlin_examples}
\end{figure}

We evaluate LocusCT on Merlin \citep{merlin}, a public dataset of approximately 25K abdominal CT exams paired with their original radiology reports. To build a visual grounding test set, we extract finding sentences using the same procedure as our labeling pipeline: Qwen3.8-27B \citep{qwen38} identifies each positive finding in a report and assigns it to a finding category. We restrict candidates to sentences containing a measurement, which tend to describe discrete, localizable findings, and randomly sample 6 findings from each of the 13 LocusBench-ED categories, yielding 78 findings. A radiologist scored LocusCT's predicted segmentation for each finding on a 5-point scale: 5 indicates that clinical measurements can be extracted directly from the segmentation, 4 that they can be extracted after minor edits, 3 that the segmentation localizes the finding but is insufficient for measurement, 2 that the segmentation is incorrect, but near the object of interest, and 1 that the segmentation is absent or far from the finding. LocusCT achieves a mean score of 3.65, with 80.8\% (63/78) of findings scoring 3 or higher and 62.8\% (49/78) scoring 4 or higher. Of the 11 findings scored 1, 8 had no predicted mask and 3 were segmented at an incorrect location. Figure~\ref{fig:merlin_examples} shows example LocusCT outputs on Merlin.


\begin{table}[t]
\centering
\caption{Ablations of LocusCT on LocusBench-Onc and LocusBench-ED. All models use the LocusCT architecture. \textit{Data} is the fraction of the full training set used; \textit{Ch.} is the number of CT window input channels (3 = soft tissue, lung, bone). The last row is our released model.}
\label{tab:ablations}
\small
\begin{tabular}{llc cc cc}
\toprule
 & & & \multicolumn{2}{c}{LocusBench-Onc} & \multicolumn{2}{c}{LocusBench-ED} \\
\cmidrule(lr){4-5} \cmidrule(lr){6-7}
Ablation & Data & Ch. & Dice  & Hit Rate & Dice  & Hit Rate  \\
\midrule
\multirow{2}{*}{Data scale}
 & 10\% & 3 & 0.267 & 0.458 & 0.282 & 0.562 \\
 & 30\% & 3 & 0.382 & 0.550 & 0.409 & 0.681 \\
\midrule
Input channels
 & 30\% & 1 & 0.353 & 0.575 & 0.373 & 0.654 \\
\midrule
Training source
 & Curated only & 3 & 0.391 & 0.596 & 0.386 & 0.658 \\
\midrule
\rowcolor{gray!15}
LocusCT (full) & 100\% & 3 & 0.495 & 0.725 & 0.490 & 0.773 \\
\bottomrule
\end{tabular}
\end{table}

\section{Discussion}

To our knowledge, this work is the first to use routine PACS annotations to create a large-scale vision-language dataset for 3D CT grounding. Trained on this weakly labeled data, LocusCT sets a new standard for grounding abdominal CT findings, achieving hit rates above 0.7 and Dice scores of approximately 0.5 on LocusBench, where existing grounding and text-promptable segmentation models perform poorly. Our scaling results further suggest that larger training sets could continue to improve performance. Finally, LocusBench provides the first expert-verified benchmarks for grounding abdominal findings, extending evaluation beyond structures covered by existing datasets.

Abdominal CT grounding is challenging because the abdomen contains many densely packed organs, while findings vary widely in appearance, location, and boundary definition. Some findings are small and well defined (e.g., kidney stones), whereas others have diffuse or poorly defined boundaries (e.g., diverticulitis). Existing grounding datasets primarily emphasize organ-level localization, whereas our pipeline targets clinically significant pathologies, including findings such as appendicitis and diverticulitis that are poorly represented in existing benchmarks. Clinical report text can also be ambiguous, with descriptions varying in specificity and anatomical detail.

This work has several limitations. First, all training data and LocusBench come from a single institution, so both the model and the benchmark may reflect site-specific scanners, acquisition protocols, patient populations, and reporting styles. Our external evaluation on Merlin offers some initial evidence of generalization, but more testing is needed. A multi-institutional grounding benchmark would provide a more rigorous test. Second, the pipeline relies in part on radiologists dictating slice numbers in their reports, which we use to link report phrases to arrow annotations. Although this is common practice, it is not universal, and sites where slice numbers are not routinely reported would yield fewer annotations. Third, LocusBench covers oncology findings and 13 ED categories. These include common and clinically important findings that existing datasets capture poorly, but they are not an exhaustive sample of the findings described in abdominal CT reports. Finally, each training and evaluation phrase describes a single positive finding, so we have not evaluated how LocusCT handles sentences with multiple or negated findings.

\section{Conclusion}
In this work, we show that grayscale softcopy presentation state (GSPS) annotations, which radiologists routinely create during clinical reading, can be turned into a large-scale source of weak supervision for 3D visual grounding. Using these annotations, we construct over 100K phrase--mask--volume triplets and train LocusCT, a 3D visual grounding model for CT. We also introduce LocusBench, a radiologist-reviewed benchmark of oncology and emergency-department findings. On LocusBench, LocusCT localizes a wide range of clinical findings described in free-text report phrases, and its performance on the external Merlin dataset indicates that it generalizes beyond the institution where it was trained. Together, these results are a step toward interactive reporting, in which findings are automatically localized in the image as the radiologist describes them.

\subsection*{AI use statement}

In this work, we used generative AI tools for the following tasks with required disclosure.
(1) \textit{Cleaning and reformatting data and qualitative data analysis:} as a core component
of our weak-labeling pipeline, an open-weight LLM (Qwen3.8-27B) extracts single-finding
sentences, matching metadata (series/slice numbers, measurements, slice positions), and finding,
organ, and region categories from radiology reports (Section~3.3; full prompts in
Appendix~A.2). This model was run locally within the institution's secure computing
environment. Its outputs were evaluated through radiologist review of the LocusBench cases
(Section~4). (2) \textit{Implementing methods:} Claude (Claude Code) and ChatGPT (Codex) assisted
in writing code for the data pipeline and model training. (3) \textit{Interpreting results:}
these tools were used to help analyze and describe experimental results.

We have not used generative AI tools to generate synthetic datasets, propose or refine
hypotheses, design research methodology or experiments, or assist with translation. Developing
theoretical models, formulating mathematical claims, and writing proofs are not applicable to
this work.

Additionally, we used generative AI tools for tasks with recommended disclosure: drafting and
editing paper text, creating and editing figures, searching for and summarizing related
literature, and formatting references and LaTeX.

We have reviewed all AI-assisted work. AI-assisted code was reviewed and tested by the authors,
and pipeline outputs were validated against radiologist review. All reported numbers were
computed by the authors' evaluation scripts and checked against raw outputs, and all
interpretations of results were verified by the authors. Literature identified or summarized
with AI assistance was checked against the original sources. No protected health information
was provided to commercial AI services; Claude and ChatGPT were used only with code, aggregate
results, and paper text. We take responsibility for the final content of this work, including
text, claims, or artifacts produced with the aid of generative AI.

\subsection*{Ethics statement}

This retrospective study was approved by the authors' institutional review board
(protocol number withheld for anonymous review), which waived the requirement for
informed consent. All imaging data, clinical annotations, and radiology reports were
accessed within the institution's secure computing environment under a HIPAA-compliant
data use protocol. Report sentence extraction and categorization were performed with an
open-weight LLM (Qwen3.8-27B) run locally, so no protected health information (PHI) was
transmitted to external services. The full weakly labeled training dataset will not be
released, in accordance with institutional data governance policies. The LocusBench images,
masks, and report phrases selected for public release were de-identified, with dates, names, and addresses replaced by placeholder tokens and report text reviewed for residual PHI prior to release. External evaluation used the publicly available Merlin dataset under its original terms of use.

LocusCT is a research model and is not intended for clinical use. Because training and
benchmark data come from a single institution, performance may not transfer to other
scanners, protocols, reporting styles, or patient populations, and we have not evaluated
performance across demographic subgroups. Our weak labels also inherit the choices of
annotating radiologists, who mark findings they consider clinically important; findings
that are routinely left unannotated may be underrepresented. We therefore caution against
deploying grounding models of this kind without prospective, multi-site validation and
human oversight. The authors declare no competing interests.

\subsection*{Reproducibility statement}

We have taken several steps to support reproducibility. Anonymized training and inference code
for LocusCT, including the configuration files that specify all training hyperparameters, is
provided in the supplementary material. Preprocessing (resampling, cropping, and three-channel
CT windowing) and the architecture and training setup are described in Section~\ref{sec:training}. The
weak-labeling pipeline is described in Section~\ref{sec:weak_labeling}: sentence extraction and GSPS matching rules
are in Section~\ref{sec:matching}, 3D mask generation with SAM2CT is in Section~\ref{sec:segmentation},
per-stage data yields are in Appendix~\ref{app:pipeline_details}, and the complete LLM prompts are in Appendix~\ref{app:prompts}.
LocusBench construction and the radiologist review protocol are detailed in Section~\ref{sec:locusbench}, and the
external evaluation on the publicly available Merlin dataset, including the sampling procedure
and scoring rubric, is described in Section~\ref{sec:merlin} and Appendix~\ref{app:merlin}. Upon acceptance, we will
publicly release LocusBench (CT volumes, radiologist-verified masks, and report phrases), the
LocusCT model weights, and the training and inference code, so that others can evaluate on the
same benchmark and compare directly against our model. The full weakly labeled training set
cannot be shared because of institutional data governance policies, but the pipeline
description and prompts are intended to allow other institutions with GSPS-annotated archives to
build comparable datasets.

\bibliography{iclr2027_conference}
\bibliographystyle{iclr2027_conference}

\newpage
\appendix
\section{Appendix}

\subsection{Pipeline Details}
\label{app:pipeline_details}

This section reports how much data remained at each stage of our pipeline.
Table~\ref{tab:sentence_matching} summarizes the report-level matching. We
consider a sentence potentially matchable to a GSPS object if it contains a
series/slice reference, a slice position (SP) value, or a measurement. Most sentences
with a series/slice reference or SP value also contain a measurement. Overall,
84.8\% of reports contained at least one matchable sentence, and this rate was
consistent across cohorts (82.3--86.9\%).

Table~\ref{tab:gsps_pooled} breaks down the GSPS annotations by object type and
shows how many were matched to a report sentence and how many were ultimately
converted into 3D segmentations. Some ellipses and less common object types
(text, angle, and multi-point annotations) could be matched to sentences by
slice position. However, we did not convert them into segmentations because
there was no clear way to interpret them consistently. As a result, none of
these object types appear in the final dataset. Our pipeline produced 109K phrase–mask–volume triplets. We excluded approximately 4K triplets from patients who appear in LocusBench to prevent patient-level overlap between training and evaluation, leaving 105K triplets for training and validation.

Table~\ref{tab:gsps_by_cohort} shows the final number of annotations converted to 3D segmentations
from each cohort. Single-line annotations are by far the most common converted annotation in
every cohort. Across cohorts, only 33--40\% of GSPS annotations were converted,
so a large share of radiologist annotations remains unused. Future work could
recover more of this signal, for example by developing consistent
interpretations for ellipse annotations or by relaxing the sentence-matching
criteria.
\begin{table}[htbp]
\centering
\small
\newcommand*{\share}[1]{{\scriptsize\color{black!55}(#1\%)}}
\setlength{\tabcolsep}{3pt}
\caption{Report sentence matching by cohort. Cohorts are disjoint; \emph{All} is their sum.
Gray percentages give each count as a share of its parent: total reports, matchable sentences,
or current-measurement sentences. Match types are not mutually exclusive, so a sentence may be
counted under more than one type.}
\label{tab:sentence_matching}
\begin{tabular}{@{}l r@{\,}l r@{\,}l r@{\,}l @{\hspace{14pt}} r@{\,}l@{}}
\toprule
 & \multicolumn{6}{c}{Cohort} & \multicolumn{2}{c}{} \\
\cmidrule(lr){2-7}
 & \multicolumn{2}{c}{Onc} & \multicolumn{2}{c}{ED} & \multicolumn{2}{c}{OED} & \multicolumn{2}{c}{\textbf{All}} \\
\midrule
\multicolumn{9}{@{}l}{\textit{Report coverage}} \\[1pt]
Total reports                          & 25,987 &               & 31,351 &               & 33,803 &               & 91,141  & \\
\quad with $\geq$1 matchable sentence  & 22,210 & \share{85.5}  & 27,234 & \share{86.9}  & 27,808 & \share{82.3}  & 77,252  & \share{84.8} \\
Matchable sentences                    & 61,246 &               & 55,050 &               & 83,489 &               & 199,785 & \\
\quad per matchable report             & 2.76   &               & 2.02   &               & 3.00   &               & 2.59    & \\
\midrule
\multicolumn{9}{@{}l}{\textit{Matchable sentences by match type}} \\[1pt]
Series/slice reference                 & 21,958 & \share{35.9}  & 16,014 & \share{29.1}  & 22,853 & \share{27.4}  & 60,825  & \share{30.4} \\
Slice position (SP)                    & 13,675 & \share{22.3}  & 273    & \share{0.5}   & 771    & \share{0.9}   & 14,719  & \share{7.4}  \\
Current measurement                    & 51,808 & \share{84.6}  & 46,766 & \share{85.0}  & 73,856 & \share{88.5}  & 172,430 & \share{86.3} \\
\quad $n=1$                            & 31,902 & \share{61.6}  & 31,780 & \share{68.0}  & 46,678 & \share{63.2}  & 110,360 & \share{64.0} \\
\quad $n=2$                            & 17,027 & \share{32.9}  & 9,875  & \share{21.1}  & 20,692 & \share{28.0}  & 47,594  & \share{27.6} \\
\quad $n=3$                            & 2,223  & \share{4.3}   & 4,705  & \share{10.1}  & 5,752  & \share{7.8}   & 12,680  & \share{7.4}  \\
\bottomrule
\end{tabular}
\end{table}

\begin{table}[htbp]
\centering
\small
\newcommand*{\share}[1]{{\scriptsize\color{black!55}(#1\%)}}
\setlength{\tabcolsep}{3pt}
\caption{GSPS annotations by type, pooled across cohorts. Gray percentages are shares of all
annotations of that type. \emph{Other} combines Text, Angle, and NPoint annotations. Ellipse and
other annotations were not converted.}
\label{tab:gsps_pooled}
\begin{tabular}{@{}l @{\hspace{10pt}} r @{\hspace{14pt}} r@{\,}l @{\hspace{14pt}} r@{\,}l@{}}
\toprule
 & & \multicolumn{2}{c}{Matched to} & \multicolumn{2}{c}{Converted to} \\
Annotation type & Total & \multicolumn{2}{c}{sentence} & \multicolumn{2}{c}{3D segmentation} \\
\midrule
Total                     & 302,868 & 129,134 & \share{42.6} & 109,401 & \share{36.1} \\
\quad Unidimensional line & 117,118 & 72,884  & \share{62.2} & 63,184  & \share{53.9} \\
\quad Arrow               & 112,388 & 29,958  & \share{26.7} & 25,106  & \share{22.3} \\
\quad Bidimensional lines & 34,303  & 24,685  & \share{72.0} & 21,111  & \share{61.5} \\
\quad Ellipse             & 37,662  & 1,504   & \share{4.0}  & ---     &              \\
\quad Other               & 1,397   & 103     & \share{7.4}  & ---     &              \\
\bottomrule
\end{tabular}
\end{table}

\begin{table}[htbp]
\centering
\small
\newcommand*{\share}[1]{{\scriptsize\color{black!55}(#1\%)}}
\setlength{\tabcolsep}{3pt}
\caption{GSPS annotations converted to 3D segmentations after sentence matching, by cohort and
annotation type. Cohorts are disjoint; \emph{All} is their sum. Gray percentages are conversion
rates: converted annotations as a share of all annotations of that type. Ellipse and other
annotations were not converted.}
\label{tab:gsps_by_cohort}
\begin{tabular}{@{}l r@{\,}l r@{\,}l r@{\,}l @{\hspace{14pt}} r@{\,}l@{}}
\toprule
 & \multicolumn{6}{c}{Cohort} & \multicolumn{2}{c}{} \\
\cmidrule(lr){2-7}
 & \multicolumn{2}{c}{Onc} & \multicolumn{2}{c}{ED} & \multicolumn{2}{c}{OED} & \multicolumn{2}{c}{\textbf{All}} \\
\midrule
\multicolumn{9}{@{}l}{\textit{All GSPS annotations}} \\[1pt]
Total                        & 88,903 &               & 109,088 &               & 104,877 &               & 302,868 & \\
\quad Unidimensional line    & 34,855 &               & 38,246  &               & 44,017  &               & 117,118 & \\
\quad Arrow                  & 36,109 &               & 44,177  &               & 32,102  &               & 112,388 & \\
\quad Bidimensional lines    & 13,095 &               & 9,317   &               & 11,891  &               & 34,303  & \\
\quad Ellipse                & 4,286  &               & 17,100  &               & 16,276  &               & 37,662  & \\
\quad Other                  & 558    &               & 248     &               & 591     &               & 1,397   & \\
\midrule
\multicolumn{9}{@{}l}{\textit{Converted to 3D segmentations}} \\[1pt]
Total                        & 31,491 & \share{35.4}  & 36,178  & \share{33.2}  & 41,732  & \share{39.8}  & 109,401 & \share{36.1} \\
\quad Unidimensional line    & 16,619 & \share{47.7}  & 21,872  & \share{57.2}  & 24,693  & \share{56.1}  & 63,184  & \share{53.9} \\
\quad Arrow                  & 7,417  & \share{20.5}  & 8,303   & \share{18.8}  & 9,386   & \share{29.2}  & 25,106  & \share{22.3} \\
\quad Bidimensional lines    & 7,455  & \share{56.9}  & 6,003   & \share{64.4}  & 7,653   & \share{64.4}  & 21,111  & \share{61.5} \\
\bottomrule
\end{tabular}
\end{table}

\begin{figure}[H]
    \centering
    \includegraphics[width=\linewidth]{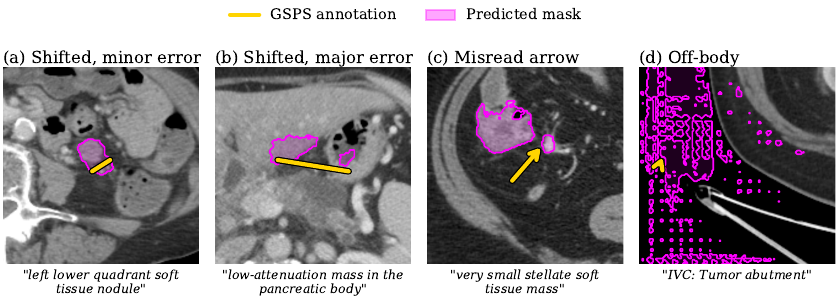}
    \caption{\textbf{Representative GSPS to 3D segmentation failure modes.} Yellow marks show the radiologist's GSPS
annotation, magenta shows the predicted mask, and the matched report sentence (partial)
appears below each panel. \textbf{(a)}~\emph{Shifted, minor error}: the mask covers the
correct nodule but is offset slightly from the annotation. \textbf{(b)}~\emph{Shifted,
major error}: the mask lands on nearby tissue and largely misses the annotated
pancreatic mass. \textbf{(c)}~\emph{Misread arrow}: instead of the small lesion at the
arrow tip, the mask segments a larger adjacent structure. \textbf{(d)}~\emph{Off-body}:
the annotation falls outside the patient, and the mask fragments across
background noise.}
    \label{fig:gsps_failure_cases}
\end{figure}

\subsubsection{GSPS Failure Cases}
GSPS objects are a valuable source of supervision, but radiologists have no
standard convention for placing them, and this variability causes errors when
converting annotations into masks. Fig.~\ref{fig:gsps_failure_cases} shows four
of the most common failure cases we encountered.

The most frequent issue is spatial shift. Measurement lines usually lie directly
over the finding they measure, but radiologists also commonly measure in the
space next to a finding, or measure on top of it and then drag the line aside so
the finding stays visible. The effect of this shift depends on its size. Small
shifts produce minor errors, where the resulting mask still captures most of the
finding (Fig.~\ref{fig:gsps_failure_cases}a). Larger shifts can cause SAM2CT \citep{SAM2CT} to
segment an entirely different structure (Fig.~\ref{fig:gsps_failure_cases}b).

An extreme form of shift occurs when the annotation is moved completely off the
body (Fig.~\ref{fig:gsps_failure_cases}d). These cases produce meaningless masks,
but they are generally easy to filter out, either by checking the quality of the
resulting mask or by detecting that the annotation lies outside the body, for
example with TotalSegmentator~\citep{total_segmentator}.

A separate failure mode occurs when SAM2CT misinterprets the annotation itself
and produces a mask that is much too large
(Fig.~\ref{fig:gsps_failure_cases}c). This is especially common for arrows,
which are the most ambiguous GSPS object type: an arrow indicates a location but
not the extent of the finding.

Despite these errors, our results show that the pipeline produces enough
high-quality annotations to outweigh the cases where it produces poor
segmentations. Reducing these failure cases remains an important direction for
future work.

\newpage

\subsection{LLM Prompts}
\label{app:prompts}

\subsubsection{Finding Identification Prompt}
Below is the prompt for identifying all findings within a report. We only preserve abnormal findings for our dataset.
\begin{lstlisting}[style=prompt]
## Task
You are given the FINDINGS section of a CT report. Reproduce it EXACTLY,
character for character, but wrap each finding in an inline tag so it can
be identified later. Do not extract findings into a separate list -- the
whole original text comes back, just with tags inserted around findings.
Reply with JSON and nothing else.

Tags: <ABN_n> for abnormal findings, <NORM_n> for normal findings, <NEG_n>
for negative findings. n starts at 1 and counts up separately within each
category, in the order findings appear -- the first abnormal finding is
<ABN_1>, the second is <ABN_2>, and so on, independent of how NORM/NEG are
numbered. Every opening tag needs its matching closing tag right after the
finding, e.g. <ABN_1>Small right pleural effusion.</ABN_1>

## The most important rule
Check every sentence for a negation -- "no" or "without" ruling something
out. If it has one, tag it NEG -- always, even when the sentence sounds
like good news. Never tag a sentence containing "no" or "without" as NORM
just because it's reassuring.
  "No small bowel obstruction." -> NEG.
  "No ascites." -> NEG.
  "PANCREAS: Normal." -> not NEG (no negation word at all).
EXCEPTION: if "no"/"without" describes the SAME problem the sentence is
otherwise reporting -- "no significant change in the mass" (the mass is
still there), "steatosis without focal lesion" (steatosis is still the
finding) -- tag the whole thing ABN or NORM instead. This exception is
rare. When in doubt, "no"/"without" means NEG.

## Choosing ABN vs NORM (only for findings with no negation)
- ABN: describes a real finding that IS present -- a mass, cyst, stone,
  thickening, nodule, degenerative change, etc.
- NORM: says a structure IS normal -- "Normal.", "Unremarkable.",
  "Clear.", "patent.", "within normal limits."

## What NOT to tag
Leave untagged (do not wrap in any tag): section headers ("PANCREAS:",
"LIVER:"), instructions to see another report, comments about image
quality or technique, and "organ is present"/"surgically absent"
statements. This text stays in the output exactly as it was, just with no
tag around it.

## Tagging a span
- One tag per finding. If a sentence names two or more separate findings,
  give each its own tag. If consecutive sentences keep describing the SAME
  finding, put them both inside ONE tag.
- Tags wrap only the finding's own words -- never include a section header
  inside a tag, even when the header is directly in front of the finding.
  The header just stays where it is, untagged, right before the tag.
- Copy the text inside every tag exactly as written -- same words,
  punctuation, capitalization, spacing, [DATE] tags. Never paraphrase or
  fix typos.

## Output
Return ONLY this JSON, nothing else -- no explanation, no markdown fences:
{"annotated_text": "..."}
The value must reconstruct to the exact original text if you strip every
tag out of it -- nothing outside a tag may be changed, added, or removed.

\end{lstlisting}

\newpage
\subsubsection{Finding Categorization Prompt}

Below is the prompt for extracting finding, organ, and region information from our sentences.
\begin{lstlisting}[style=prompt]
## Task
You are given a short piece of text describing exactly one confirmed abnormal radiology finding from a CT report (this finding already has a matching image segmentation mask, so it IS a real finding -- never decide that it isn't one, and never refuse to answer). Assign it three labels by choosing the best-fitting entry from three running category banks below:
  1. finding -- what KIND of abnormality this is (e.g. Cyst, Mass, Kidney Stone).
  2. organ   -- what organ or anatomic structure the finding is IN or ON.
  3. region  -- what broad body region the finding is located in.
Reply with JSON and nothing else.

## Reading the input
You will be given the input in one of two forms:
- IN-CONTEXT -- a longer excerpt of surrounding report text, with the exact
  span describing the finding of interest wrapped in <REF></REF>. Text
  outside <REF></REF> is background only (organ-section headers, nearby
  sentences) -- it is NEVER itself the target, even if it also names an
  abnormal-sounding structure.
- SENTENCE -- just the finding's own sentence(s), with no surrounding
  context given. Treat the entire given text as the target.
Only the <REF>...</REF> span (or, when there is no IN-CONTEXT, the whole
SENTENCE text) can contain the target finding.

## When the target span names more than one candidate entity
Sometimes the target text itself names two or more anatomical entities or findings (e.g. comparing today's finding to an old, unrelated one, or mentioning a nearby incidental structure in the same breath). Pick the one that is accompanied, in that same text, by:
  - a measurement (e.g. "1.2 cm", "8 mm"), or
  - a slice/series/image reference (e.g. "series 3, image 40"), or
  - a "SP" reference (also sometimes "TP").
That accompanied entity is the actual target -- never a structure that is only named incidentally (for comparison, as a distractor, or in passing), with no measurement/slice/SP/TP of its own attached to it. If only one entity is named, there is no ambiguity -- just use it. Decide from the given text alone -- never use any information beyond it (no image or mask file names, no other fields).

## The category banks -- reuse first, invent only when truly necessary
Below are the CURRENT entries in each of the three banks. Categories added earlier in this run are already permanent members, exactly like the original seed entries -- treat them identically; there is no difference in status between a seed entry and one added later.

Existing FINDING categories: {finding_bank}
Existing ORGAN categories: {organ_bank}
Existing REGION categories: {region_bank}

For EACH of finding / organ / region, follow this priority order:
  1. If ANY existing bank entry is a reasonable fit -- even an imperfect
     one, even if more than one entry could arguably fit -- choose the
     single best-fitting existing entry. Do not refuse or hedge.
  2. Only if NOTHING currently in that bank is a reasonable fit for what
     the text actually describes, answer "OTHER" for that field and give
     a new category name in the matching new_* field.
Bias strongly toward option 1. A new category becomes a PERMANENT addition that every later sentence in this run will also be offered -- so before proposing one, check hard whether it is really a distinct clinical concept, or just a rephrasing, synonym, abbreviation, or plural of something already in the bank (e.g. "renal calculus", "urolithiasis", and "stone" are all just Kidney Stone; "growth" or "tumor" is usually just Mass or Lesion depending on context). If it is the same concept under another name, reuse the existing entry instead of proposing it. When you do propose a new category (field = "OTHER" and the matching new_* field is filled in), name it the way existing bank entries are named: a concise, Title Case, singular clinical noun phrase (e.g. "Splenomegaly", "Small Bowel Obstruction") -- not a full sentence, not lowercase, not plural, and never the literal word "Other". organ starts with NO entries -- the earliest sentences in a run will often need "OTHER" for organ. Once an organ (e.g. "Kidney", "Liver") is in the bank, treat it exactly like a seed category and prefer reusing it.

## Output
Reply with ONLY this JSON object, nothing else -- no explanation, no
markdown fences:
{{"finding": "...", "new_finding": null, "organ": "...", "new_organ": null, "region": "...", "new_region": null}}
- finding / organ / region: must be exactly one of that field's current
  bank entries listed above, OR the literal string "OTHER".
- new_finding / new_organ / new_region: null whenever the matching field
  above is NOT "OTHER". When the matching field IS "OTHER", give the new
  category name here (non-empty, Title Case, singular) -- never leave it
  null or empty in that case.

\end{lstlisting}
\subsection{LocusBench Results}
\label{app:per_category}
\begin{figure}[H]
    \centering
    \includegraphics[width=0.5\linewidth]{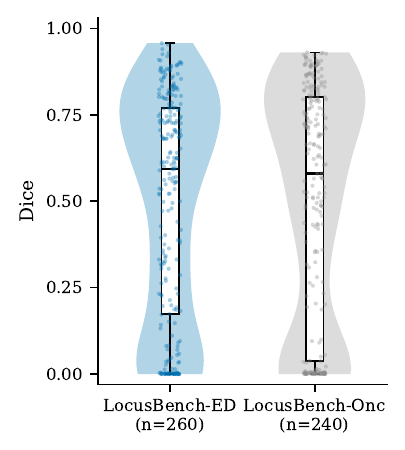}
    \caption{\textbf{Dice score distributions for LocusCT on LocusBench-ED ($n=260$) and
LocusBench-Onc ($n=240$)}. The two splits
have similar medians (0.59 ED, 0.58 Onc), but both are bimodal: most cases
cluster at high Dice, with a second group near zero corresponding to complete localization
failures. LocusBench-Onc has more of these failures, so its lower quartile falls near zero
(0.04, vs. 0.17 for ED).}
    \label{fig:boxplot_overall}
\end{figure}

\begin{figure}[H]
    \centering
    \includegraphics[width=\linewidth]{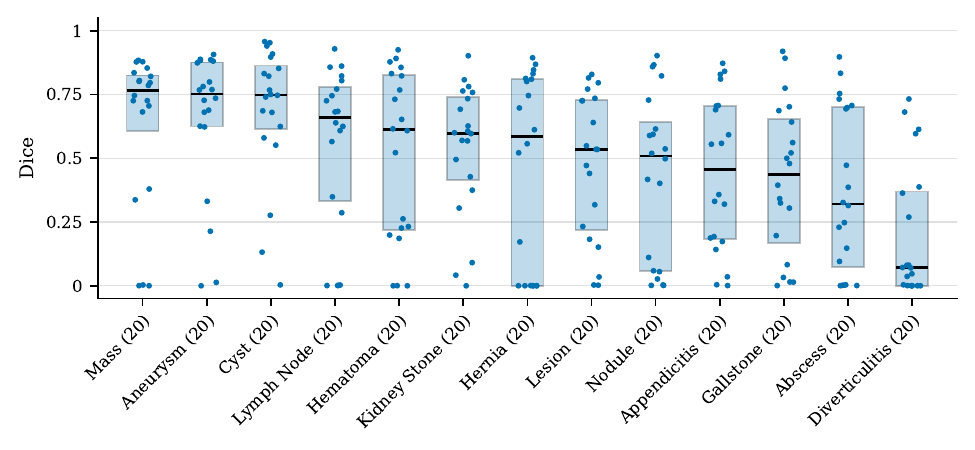}
    \caption{\textbf{Per-category Dice scores for LocusCT on LocusBench-ED, sorted by median
Dice.} Each category contains 20 cases (shown in parentheses). Points are
individual cases, boxes span the interquartile range, and black lines mark the
median. Many categories are bimodal: most cases are segmented well, but some
score near zero, meaning the model failed to localize the target finding at all
rather than producing a poor boundary. Performance is strongest for mass,
aneurysm, and cyst and weakest for diverticulitis and abscess.}
    \label{fig:boxplot_ed_by_category}
\end{figure}

This section presents more detailed results for LocusCT on LocusBench.
Fig.~\ref{fig:boxplot_overall} shows the distribution of Dice scores across all
of LocusBench-ED and LocusBench-Onc. LocusCT achieves nearly the same median Dice
on both splits (0.59 and 0.58). Both distributions are also bimodal: most cases
either reach a high Dice score or fall close to zero, with relatively few in
between. This suggests that when the model localizes a finding correctly, it
usually produces a good segmentation, and that most errors come from failing to
localize the finding at all. This supports our motivation that localization,
rather than segmentation, is the central challenge of 3D visual grounding in CT.

Fig.~\ref{fig:boxplot_ed_by_category} breaks down the LocusBench-ED results
across its 13 categories. Many individual categories show the same bimodal
pattern. Mass, aneurysm, and cyst have the highest median Dice scores, while
abscess and diverticulitis perform worst, with diverticulitis reaching a median
of only 0.07. Overall, 9 of the 13 categories have a median Dice above 0.5.

\subsection{Merlin Results}
\label{app:merlin}
\begin{figure}[H]
    \centering
    \includegraphics[width=\linewidth]{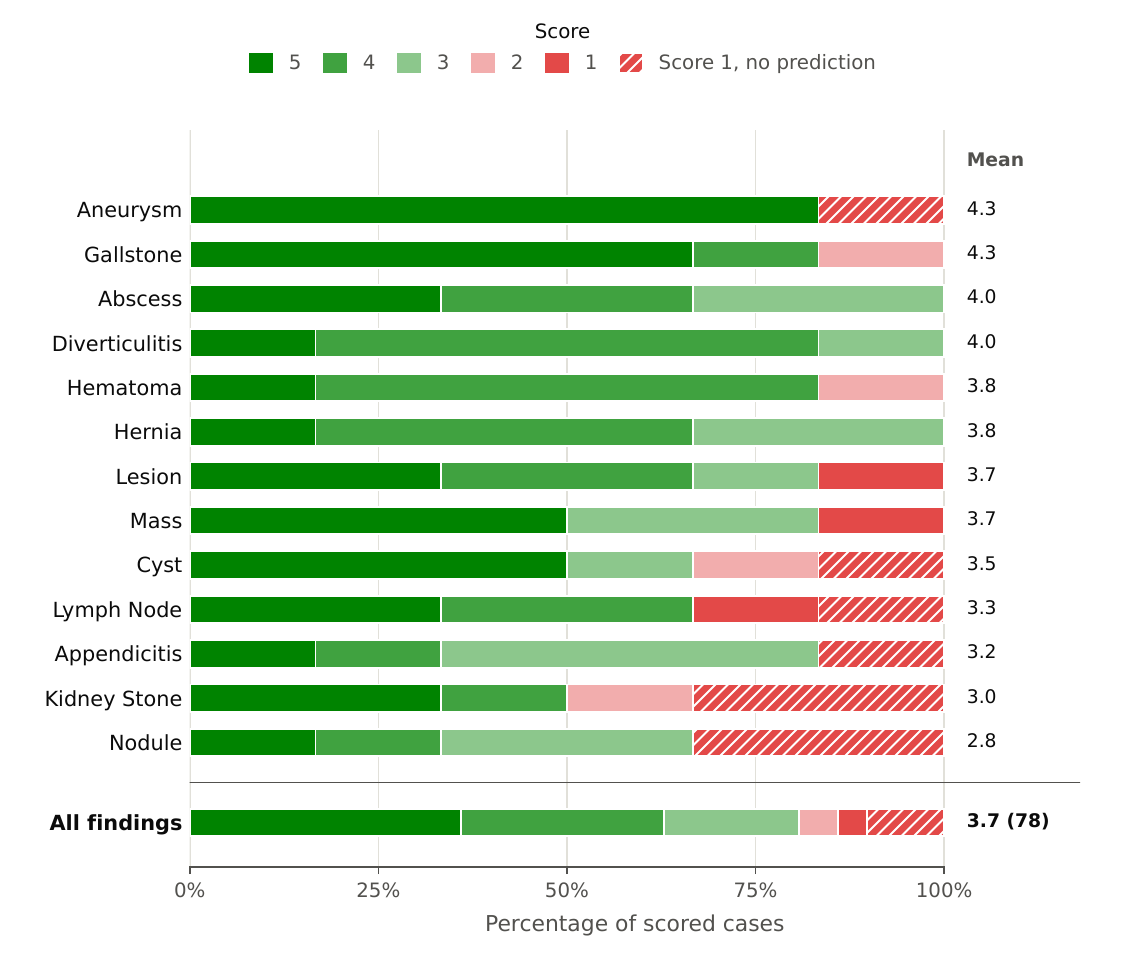}
    \caption{\textbf{Reader study of LocusCT grounding on the external Merlin dataset.}
For each of the 13 LocusBench-ED finding types, six abdominal CT scans from the Merlin
dataset were grounded with LocusCT using the corresponding report phrase, and each
predicted mask was rated by a board-certified radiologist on a
5-point scale (1 = Missing/Not near the object of interest; 2= Incorrect, but near the object of interest; 3 = Useful for localization, but not for information extraction; 4 = Useful, only minor edits required for clinical measurements; 5 = Very useful: could be used for clinical measurements).
Bars show the distribution of ratings per finding type, sorted by mean rating. Hatched
segments mark cases where the model produced no prediction; these are assigned a score
of 1 and included in the means. The bottom row pools all 78 cases.}
    \label{fig:merlin_results}
\end{figure}

This section presents a more detailed breakdown of our external evaluation on
Merlin, shown in Fig.~\ref{fig:merlin_results}. Over 75\% of the evaluated
findings were judged useful for localization, which is close to LocusCT's hit
rate of 0.773 on LocusBench-ED. This suggests that the model's localization
ability generalizes to an external dataset.

Some categories performed noticeably worse on Merlin than on LocusBench-ED. For
example, kidney stone performed poorly here despite doing well on LocusBench-ED.
However, each category contains only six cases, so per-category results should
be interpreted with caution.

These results also suggest that, when LocusCT fails, it tends to output no mask
rather than an incorrect one. This is the preferable failure mode in both
research and clinical settings, since an empty output is easy to detect and
flag, whereas an incorrect mask could mislead a downstream user.





\end{document}